# Are Lexicon-Based Tools Still the Gold Standard for Valence Analysis in Low-Resource Flemish?

Ratna Kandala, Katie Hoemann

Department of Psychology, University of Kansas, USA

n038k926@ku.edu, hoemann@ku.edu

## ABSTRACT

Understanding the nuances in everyday language is pivotal for advancements in computational linguistics and emotions research. Traditional lexicon-based tools such as Linguistic Inquiry and Word Count (LIWC) (Pennebaker et al., 2007) and Pattern (De Smedt & Daelemans, 2012) have long served as foundational instruments in this domain. LIWC is the most extensively validated word-count-based text analysis tool in the social sciences and Pattern is an open-source Python library offering functionalities for Natural Language Processing (NLP). However, everyday language is inherently spontaneous, richly expressive, and deeply context dependent. To explore the capabilities of Large Language Models (LLMs) in capturing the valences of daily narratives in Flemish (Belgian Dutch), a low-resource language, we first conducted a study involving approximately 25,000 textual responses from 102 Dutch-speaking participants. Each participant provided narratives prompted by the question, "What is happening right now and how do you feel about it?", accompanied by self-assessed valence ratings on a continuous scale from –50 to +50. We then assessed the performance of three Dutch-specific LLMs ChocoLlama-8B-Instruct, Reynaerde-7B-chat, and GEITje-7B-ultra in predicting these valence scores, and compared their outputs to those generated by LIWC and Pattern. Our findings indicate that, despite advancements in LLM architectures, these Dutch-tuned models currently fall short in accurately capturing the emotional valence present in spontaneous, real-world narratives. This study underscores the imperative for developing culturally and linguistically tailored models/NLP tools that can adeptly handle the complexities of natural language use. Enhancing automated valence analysis is not only pivotal for advancing computational methodologies but also holds significant promise for enriching psychological research with ecologically valid insights into human daily experiences. We advocate for increased efforts in creating comprehensive datasets and finetuning LLMs for low-resource languages like Flemish, aiming to bridge the gap between computational linguistics and emotion research.

## 1. Introduction

The increasing use of Large Language Models (LLMs) in everyday applications has raised important questions about their ability to understand and reflect emotional nuances in natural language communication. As users increasingly seek empathetic interactions with LLMs perceiving them as trustworthy and emotionally resonant agents, the ability to decode valence (emotional pleasantness/unpleasantness, communicated via language) becomes critical. Yet, current LLMs lack the psychological mechanisms to generate genuine empathy, as noted in a recent Nature Intelligence editorial (Shteynberg, 2024), underscoring the urgency of culturally and linguistically tailored solutions. This can have long-term effects on how AI systems are experienced by humans. This gap is especially pronounced in low-resource languages like Flemish (Belgian-Dutch), where culturally nuanced valence detection is essential to develop Natural Language Processing (NLP) tools for Flemish speakers.

Lexicon-based sentiment analysis employs predefined dictionaries where each word is associated with a sentiment label such as positive, negative, or neutral or a sentiment score indicating its emotional intensity. These dictionaries enable the extraction of sentiment from text by mapping words to their corresponding emotional values. This approach is widely utilized in psychological research to examine emotional content in text and analyze verbal behavior (Pennebaker et al., 2007; Tausczik & Pennebaker, 2010). Lexicon-based methods have also been particularly useful in studying emotional references in natural language (Mohammad & Turney, 2013). Traditional lexicon-based tools like Linguistic Inquiry and Word Count (LIWC) (Pennebaker et al., 2015) have long dominated valence analysis in psychological research. These methods rely on predefined word-emotion mappings, offering interpretability. For instance, LIWC, the widely used tool for text analysis in emotion research, excels in validated dictionary categories but faces limitations in capturing spontaneous, context-rich narratives common in daily language. For example, in a study survey analyzing responses about workplace satisfaction, LIWC's predefined "PosEmo" dictionary might correctly flag words like "fulfilled," "collaborative," or "rewarding" as positive, offering interpretable insights into employee morale. However, in informal narratives like "This 'amazing' Monday has me stuck in traffic for hours - what a joy!", LIWC would misclassify "amazing" and "joy" as positive, failing to detect the sarcastic tone and contextual frustration. Similarly, another tool - Pattern's (De Smedt & Daelemans, 2012) rule-based sentiment analysis, while robust for structured text, falters with free-form expressions due to its reliance on syntactic patterns. For instance, Pattern's syntactic rules might accurately score the sentence "The product works flawlessly and is incredibly user-friendly" as positive by recognizing adjectives like "flawlessly" and "user-friendly." Yet, in a free-form narrative like "It's wild how this 'perfect' app crashes every time I need it most - truly next-level reliability," Pattern would assign a positive score based on "perfect" and "next-level," missing the irony and negative sentiment conveyed through context.

Most work in the sentiment analysis domain has been done in languages such as English, French, German, Mandarin, Spanish etc. which benefit from extensive annotated datasets. The amount of literature on English outweighs other languages in sentiment analysis (Ligthart et al., 2021). Recent advances in LLMs like GPT-4o (Martinez, 2024) have shown promise in valence prediction for English. However, their efficacy in low-resource languages remains unproven, particularly for Flemish, which exhibits lexical, grammatical, and sociolinguistic features distinct from standard (Netherlandic) Dutch. For example, Flemish incorporates regional vocabulary (e.g., "fuif" for "party"), divergent pronoun usage ("gij" vs. "jij"), and multilingual influences from French (and, like other variants of Dutch, increasingly English). These nuances challenge lexicon-based tools, risking inaccurate valence detection in Flemish texts.

This study addresses three critical gaps:

1. Linguistic Inequity: Flemish, spoken by 6.5 million in Belgium, remains underrepresented in NLP research (Meeus et al., 2024).

2. Ecological Validity: Prior work predominantly relies on social media data (e.g., tweets) or online surveys, which impose character limits and linguistic constraints that distort natural emotional expression leading to a performance decrease in sentiment analysis (Reusens, 2022; Roberts et al., 2014) in contrast to open-ended responses.

3. Methodological Rigor: Existing evaluations of LLMs for valence analysis often use external annotators, introducing biases absent in self-reported ground truth.

We present a comparison of lexicon-based tools (LIWC, Pattern) and Dutch-tuned LLMs (ChocoLlama-8B-Instruct, Reynaerde-7B-chat, GEITje-7B-ultra) using nearly 25,000 open-ended Flemish narratives collected via ambulatory assessment (where participants repeatedly report experiences in real time within their natural environments, minimizing the retrospective bias inherent to lab surveys) from 102 Dutch speakers in Belgium. Participants described their real-time experiences using self-rated valence scores (−50 to +50). Our findings reveal that even state-of-the-art Dutch LLMs underperform lexicon-based methods in aligning with human valence ratings, underscoring the need for linguistically tailored models. This gap carries profound implications for low-resource NLP studies, especially for emotion science researchers. Unlike social media data, these narratives reflect free-flowing, context-rich expressions of valence, offering ecologically valid insights into momentary well-being. As such this is an attempt to bridge computational linguistics and psychology for low-resource languages, enabling models that respect linguistic diversity. Our dataset, the first of its kind for Flemish, also paves the way for LLMs that genuinely understand cultural nuance, a prerequisite for applications in customer sentiment analysis, social science research, and personalized mental health interventions. This study underscores the societal stakes of equitable AI development: inaccurate valence detection risks alienating Flemish speakers from mental health support tools, exacerbating disparities in AI-driven care.

This paper is structured as follows. First, we situate our work within the broader literature on sentiment analysis and emotion detection, focusing on low-resource languages and methodological gaps in NLP. Next, we provide a concise overview of the tools central to this study: the lexicon-based frameworks *LIWC* and *Pattern*, and Dutch-tuned large language models. Following this, we outline our methodology, detailing the collection of nearly 25,000 Flemish daily narratives, the self-reported valence rating protocol, and the comparative evaluation framework for LLMs and lexicons. Subsequently, we present results quantifying the alignment between model predictions and human valence ratings, emphasizing coverage disparities and statistical significance. We then provide a discussion contextualizing the performance disparities between lexicon-based tools and Dutch-tuned LLMs. Finally, we conclude with practical recommendations for advancing equitable NLP tools in under-resourced linguistic contexts and propose future research directions to bridge computational and psychological inquiry.

2. Prior Work

Sentiment analysis (SA), or the detection of valence in language [(un)pleasantness], is widely applied in healthcare (Qiu et al., 2023), customer service (Ali, 2024), and misinformation detection (Liu, 2023). Traditional lexicon-based tools like LIWC (Pennebaker et al., 2015), Pattern (De Smedt & Daelemans, 2012), and VADER (Hutto & Gilbert, 2014) which map words to predefined sentiment scores have dominated psychological research due to their interpretability. However, these tools lack contextual awareness, struggle with sarcasm, multilingual code-switching, and domain-specific valence cues (Mohammad & Turney, 2013). Recent studies highlight LLMs' potential in valence detection, particularly GPT-4o for English texts (Martinez, 2023). While LLMs also show promise in psychiatry (Obradovich et al., 2024) and psychotherapy (Mishra et al., 2023), their performance in low-resource languages remains understudied. Untuned LLMs often fail to capture emotion intensity (Liu et al., 2024), and their reliance on English-centric training data limits cross-linguistic generalizability (Ligthart et al., 2021).

For languages such as Dutch and its regional variant Flemish, spoken by 6.5 million in Belgium, with distinct vocabulary (e.g., *fuif* for "party") and pronoun usage (*gij* vs. *jij*), SA research is sparse. Recently, Reusens (2022) compared lexicon methods with BERT-based models on Flemish tweets, finding transformers superior for short-text analysis. However, this work omitted LIWC, the psychology community's gold standard, and Pattern, and relied on social media data constrained by character limits and linguistic norms (Roberts et al., 2014). Most SA datasets use external annotators to label valence (Dashtipour et al., 2021; Alexandridis et al., 2021), introducing biases from third-party interpretations. Social media posts further distort emotional expression due to self-censorship and platform-specific constraints (Reusens, 2022). In contrast, ambulatory assessment - collecting real-time, open-ended narratives with self-reported valence offers ecologically valid ground truth (Pennebaker, 1997; Frisina et al., 2004). We address these gaps by: (a) Evaluating three Dutch-tuned LLMs against LIWC and Pattern on nearly 25,000 open-ended Flemish narratives with self-reported valence scores (−50 to +50). (b) Using the untranslated (to English), context-rich daily narratives to preserve linguistic authenticity (c) Providing a comparison of lexicon-based methods and LLMs for low-resource Flemish.

### 3. Lexicon-Based Text Analyses

A lexicon-based method employs a dictionary specifically created to detect sentiments, with its core being a lexicon resource characterized by accuracy and comprehensive vocabulary coverage. Within the landscape of computational linguistics tools, lexicon-based approaches are valued for their interpretability and ease of implementation. Below, we briefly discuss LIWC and Pattern.

### 3.1 LIWC

Linguistic Inquiry and Word Count (LIWC), introduced by Pennebaker et al. (2001), is among the earliest and most widely adopted frameworks. Although initially developed for English, LIWC has been adapted for Dutch and other languages (Boot et al., 2017). Its distinguishing feature is a multidimensional classification of language, mapping Dutch lexicon words across more than 90 categories, including psycholinguistic features like emotional tone (PosEmo and NegEmo) and general metrics such as word frequency and punctuation use. While the PosEmo (positive emotion) and NegEmo (negative emotion) categories are commonly leveraged for SA, LIWC's extensive categorization also supports applications in psychology (e.g., identifying markers of depression) and sociolinguistics (e.g., examining gender-related communication patterns). To estimate the valence [(un)pleasantness] of texts using LIWC, researchers calculate the percentage of dictionary-matched words within a given text. The Dutch version of LIWC 2015 (Boot et al., 2017) contains 6,614 words (of which 1227 words belong to PosEmo and 1474 words belong to NegEmo). However, as a proprietary closed-source product, LIWC lacks transparency for customization, and its methodology does not incorporate rules for detecting negation, sarcasm, or intensifiers.

### 3.2 Pattern

Pattern is an openly available Python package developed by the CLiPS Computational Linguistics group, University of Antwerp (De Smedt & Daelemans, 2012). Although it is often introduced as a web-mining library, it bundles a broad Natural Language Processing (NLP) stack: web-scrapping helpers, part-of-speech taggers, n-gram utilities, and WordNet access for six languages (English, Spanish, German, French, Italian, Dutch). Its advantages, such as easy installation, have become a convenient tool for researchers who do not want to juggle multiple specialized libraries. It also provides a dedicated sub-module for Dutch (Pattern.nl) to perform SA without any machine learning knowledge. Unlike recent studies using large language models (LLMs) for SA that demand GPU resources, fine-tuning data and coding expertise, Pattern.nl is genuinely "plug-and-play." Installing the package and calling a single function (sentiment(text)) is sufficient, significantly lowering the threshold for scholars in psychology who need scalable sentiment estimates without the need to invest in a full ML pipeline to answer their research questions. To perform SA, Pattern.nl relies on a handcrafted polarity lexicon containing 3,304 distinct Dutch lemmas, approximately 97% of which are adjectives, the part-of-speech (POS) that carries the most evaluative meaning. Each lemma is annotated with a polarity value in the interval [-1,1] and auxiliary metadata such as subjectivity strength and short glosses. The algorithm works in three major steps: chunking, chunk-level scoring, and sentence-level aggregation. In chunking, a sentence is divided into chunks (e.g. nominal, verbal, adjectival chunks). In chunk-level scoring, within each chunk, every token that appears in the lexicon contributes to its polarity. Through hand-coded rules, tokens flagged as intensifiers, diminishers or negators adjust the values accordingly. Importantly, Pattern distinguishes between degree-modifying adverbs and regular sentiment words based on POS tags. For example, 'verschrikkelijk' can act as an adjective ("horrible") or as an adverbial intensifier ("terribly nice"). In sentence-level aggregation, the final sentence polarity is the average of the regulated chunk scores. However, the lexicon was originally compiled from product-review corpora and has been reported to score texts as neutral for sentences with affective words outside the domain on product reviews, simply because the lemmas were absent from the resource.

## 4. LLMs Chosen

To evaluate valence estimation in Flemish texts, we selected three open-weight LLMs fine-tuned for Dutch: ChocoLlama-8B-Instruct, GEITje-7B, and Reynaerde-7B-Chat. These models were chosen for their architectural diversity and training methodologies, aiming to assess their capabilities in understanding and processing Flemish emotional content. The selection of open-source models was also driven by considerations for data privacy, ensuring that participant data remains secure. All checkpoints are released under the Apache-2.0 licence and load with the standard HuggingFace chat template.

**Llama-3-ChocoLlama-8B-Instruct** (Meeus et al., 2024): Built upon Meta's Llama-3-8B architecture, this model employs rank-16 low-rank adaptation (LoRA) on 32B Dutch tokens, followed by supervised fine-tuning (SFT) and direct-preference optimization (DPO) alignment. Hugging Face It adheres to a language-adaptation strategy that retains the original tokenizer, a method previously shown to enhance Dutch performance in Llama-2 variants. A dedicated evaluation set, ChocoLlama-Bench, accompanies the release.

**GEITje-7B-Ultra** (Vanroy, 2024): Based on Mistral 7B architecture, GEITje-7B-Ultra undergoes full-parameter pretraining on 10 billion Dutch tokens, followed by supervised fine-tuning (SFT) and direct preference optimization (DPO) alignment for conversational use. Hugging Face Project documentation indicates that GEITje-Ultra was built on publicly available Llama-2-7B-hf weights (Touvron et al., 2023), reflecting an open-foundation philosophy. Among the GEITje family, the Ultra variant is reported as the best overall performer.

**Reynaerde-7B-Chat** (ReBatch, 2024): Rather than further pretraining, Reynaerde-7B-Chat attaches rank-8 quantized LoRA (QLoRA) adapters to a Llama-2-7B-hf base. It is fine-tuned on $4 \times 10^5$ Dutch chat pairs and concludes with DPO alignment, producing a safe, concise dialogue style that can run on a single NVIDIA A100 GPU.

It has been demonstrated that both GEITje-7B-Ultra and Reynaerde-7B-Chat have outperformed ChocoLlama-2-7B-tokentrans-instruct in Dutch reasoning, comprehension, and writing benchmarks, though their efficacy for psychological research remains to be evaluated (Meeus et al., 2024).

## 5. Methodology

### 5.1 Data

The dataset comprises 24,854 open-ended Dutch-language text entries (medium-scale) collected longitudinally over 70 days from 102 participants (age 18–65, µ=26.47, σ=8.87) in Belgium. Participants were native Dutch speakers with smartphone access, recruited via online/posted flyers and referrals (most were native Belgians, a few were from the Netherlands and living in Belgium). Using a dedicated app, they received four daily prompts asking (in Dutch), "What is going on now or since the last prompt, and how do you feel about it?" Responses were either short text snippets (3-4 sentences) or 1-minute voice recordings, yielding a temporally structured dataset with potential for time-series or multimodal analysis (text + audio). For robustness, eligibility criteria ensured linguistic consistency (native speakers) and device reliability (smartphone ownership). Further methodological details, including ethical protocols, are provided in the Supplementary Material. The dataset can be made available to interested persons after acceptance.

### 5.2 Prompt for the LLMs

To enable sentiment analysis on this dataset, we designed a standardized prompt to elicit valence ratings from various Dutch-language large language models (LLMs):

> "You are a Dutch language expert analyzing the valence of Belgian Dutch texts. Participants responded to:
>
> 'What is going on now or since the last prompt, and how do you feel about it?'
>
> Carefully read the response of the participant: {text}. Your task is to rate its sentiment from 1 (very negative) to 7 (very positive). Return ONLY a single numerical rating enclosed in brackets, e.g. [X], with no additional text.
>
> Output Format: [number]"

The placeholders {text} are filled with the texts.

## 6. Results: Performance of LLMs versus Lexical Tools

This section presents the results of the models and their comparison with self-reported valences of the users.

### 6.1 Coverage of Valence Predictions

We first assessed the practical applicability of each method by evaluating coverage—the proportion of texts (N = 24,852) for which models generated a numeric valence score because output coverage varied substantially across models. Lexicon-based tools demonstrated near-complete coverage: LIWC and Pattern produced scores for 24,848 texts (99.9%). In stark contrast, Dutch-tuned LLMs exhibited substantial variability. ChocoLlama-8B-Instruct generated predictions for 17,378 texts (69.9%), GEITje-7B-ultra for 9,445 texts (38.0%), and Reynaerde-7B-chat for only 446 texts (1.8%). These disparities underscore a critical challenge: while lexicon tools universally processed inputs, LLMs' coverage gaps risk introducing selection bias, complicating fair performance evaluations.

### 6.2 Alignment with Self-Reported Valence

Next, we evaluated how well each model's predictions align with the user's/author's self-reported valence ratings using correlation metrics. Higher correlation meant the model's predicted sentiment closely tracks the writer's own valence ratings which range from –50 for extremely negative sentiment to +50 for extremely positive sentiment. This was done using two correlations: (a) Pearson's r, the standard measure of linear association between two continuous variables (b) Polyserial correlation, a statistical measure used to assess the relationship between an ordinal variable (i.e. the model's prediction) to a continuous variable (user's rating). All analyses were limited to the subset of texts processed by each model.

### 6.3 Model Performance Evaluation: Lexicon-Based Valence Prediction

In an analysis of valence prediction performance, the following correlation coefficients were observed for lexicon-based methods: Table 1 compares LIWC's (PosEmo and NegEmo), and Pattern.nl against the user's self-reported valence. Pattern showed the highest correlation (Pearson r = 0.31, Polyserial r = 0.31) when compared to LIWC15-PosEmo (Pearson r = 0.21, Polyserial r = 0.23) and LIWC15-NegEmo (Pearson r = -0.23, Polyserial r = -0.23). Thus, between LIWC15 and Pattern, Pattern turned out to be the robust lexicon-tool for valence prediction for these Flemish texts.

Table 1: Correlation and coverage metrics across sentiment models

| Model | Coverage (N/%) | Pearson $r$ | Polyserial $r$ |
|---|---|---|---|
| LIWC15 (posemo) | 24,848 / 99.9% | 0.21 | 0.23 |
| LIWC 15 (negemo) | 24, 848 / 99.9% | -0.23 | -0.23 |
| Pattern.nl | 24,848 / 99.9% | 0.31 | 0.31 |
| Chocollama-8B-instruct | 17,378 / 69.9% | 0.35 | 0.40 |
| GEITje-7B-ultra | 9,445 / 38% | 0.35 | 0.44 |
| Reynaerde-7B-Chat | 446 / 1.8% | 0.18 | 0.24 |

Coming to the large language models, on the 17,378 texts for which ChocoLlama-8B-Instruct returned a numeric valence score, the Pearson correlation was 0.35 and a polyserial correlation was 0.40 with the user's self-reported valence. On this same subset of texts, the lexicon baseline LIWC yielded substantially lower linear associations. LIWC positive-emotion counts correlated at r = 0.23 and LIWC-negative emotion counts at r = -0.23, and the Pattern composite score at r = 0.32. Paired significance texts/Paired samples t-test confirmed that Pattern outperformed the LIWC NegEmo feature (t = 10.57, p < 0.001). Crucially, ChocoLlama's correlation was significantly higher than Pattern's (t = 4.02, p < 0.001). Together, these results suggested ChocoLlama's stronger alignment with the author's continuous valence ratings than individual LIWC and Pattern for these subsets of texts.

Speaking of the next model, GEITje-7B-ultra, across the 9445 texts it scored, its valence predictions exhibited a Pearson correlation of r = 0.35 and a polyserial correlation of 0.44 with the user's self-reported ratings. When evaluated on this identical subset, the lexicon methods delivered weak associations: LIWC PosEmo yielded r =0.23, LIWC NegEmo r = -.26, and Pattern r =0.34. A paired comparison confirmed that Pattern's correlation significantly exceeded LIWC's negative-emotion score (t = 7.14, p < 0.001), whereas GEITje's correlation did not significantly differ from Pattern's (t = - 1.20, p = 0.23). Thus, although both GEITje and Pattern performed better than the individual LIWC dimensions, GEITje does not provide a significantly reliable improvement over the best lexicon baseline on the texts it has covered.

Now coming to Reynarrde-7B-chat, in a comparative analysis for the 446 texts it returned values for, the model achieved a Pearson correlation of 0.18 and a Polyserial correlation of 0.24 with the user self-reported valence scores in a scale from –50 to +50. These results were benchmarked against LIWC and Pattern which yielded the following correlations: LIWC-PosEmo: 0.25, LIWC-NegEmo: -0.19, Pattern: 0.28. Statistical significance of performance differences was assessed via paired t-tests (n = 446). The Pattern's method's correlation did not significantly surpass LIWC-NegEmo's (t = 1.48, p = 0.14) or Reynaerde-7B-chat's (t = 1.57, p =0.12). While Pattern exhibited nominally higher correlations than Reynaerde, the difference lacked statistical reliability at this sample size.

To summarize, ChocoLlama-8B-Instruct and GEITje-7B-ultra exhibited higher correlations (Pearson r = 0.35) than lexicon tools on their respective subsets, but their incomplete coverage (69.9% and 38.0%) limits their applicability. Lexicon methods, by contrast, achieved near-universal coverage (99.9%) with moderate correlations (Pattern: r = 0.31).

## 7. Discussion

Our study was an attempt to evaluate how well Dutch-finetuned large language models (LLMs) capture emotional valence in daily Flemish narratives compared to lexicon-based tools. While LLMs like ChocoLlama-8B-Instruct and GEITje-7B-ultra showed moderate correlations with human ratings on subsets of texts, their performance was constrained by critical limitations: incomplete coverage and possibly, a systemic domain mismatch between training data and real-world narratives. In contrast, lexicon-based tools like Pattern.nl, though less contextually nuanced, provided reliable valence estimates across the entire corpus. These findings challenge the assumption that language-specific LLMs inherently outperform lexicon methods in low-resource settings and highlight the need for culturally tailored solutions.

Here we discuss the plausible reasons for the fine-tuned LLMs on Dutch to yield the observed results for our daily narratives Flemish dataset. Most of the LLMs' general training data are based on web and one-third of all data is in English (Mattheus, 2024). For example, Llama-2's pretraining data comprises 89.7% English and only 10.3% non-English, leaving colloquial and pragmatic features of Flemish underrepresented (Li et al., 2024). GPT-3 exhibits a similar skew, with roughly 92.65% of its tokens in English, underscoring a systemic English-centric bias across major LLMs. Adaptation efforts like ChocoLlama which incorporated 32 billion Dutch token for continued pretraining demonstrated sizeable gains on Dutch benchmarks confirming that language-specific corpora are essential to overcome these blind spots (Mattheus, 2024). However, overreliance on publicly available corpora (e.g. legal documents, social media texts) imbues LLMs with a formal, technical register that rarely features colloquialism, first-person constructions, and implicit emotional cues characteristic of everyday narratives, leading to blind spots and inaccurate valence estimates in informal texts. This domain mismatch undermines LLM performance in terms of its accuracy on valence estimation simply because these models lack exposure to the lexical, syntactic, and pragmatic cues that signal emotion in everyday narratives (Blitzer et al., 2007).

In contrast, LIWC and Pattern.nl are grounded in psycholinguistic or manually curated word lists to retain sensitivity to informal affective expressions (Pennebaker et al., 2001; Gatti & van Stegeren, 2020). LIWC's validated dictionaries map words directly onto positive and negative affect, producing interpretable scores that align closely with human judgments across registers (Pennebaker et al., 2001). Notably, Pattern.nl outperformed LIWC in our study, likely due to its composite polarity scoring rather than LIWC's reliance on raw emotion-word counts.

Addressing this gap requires augmenting training data with narrative-style corpora or combining LLM outputs with lexicon-driven scores to create hybrid models better suited for daily Flemish narratives. They are not specifically developed datasets for capturing the valences. Pretrained LLM embeddings offer general distributional semantics learned from massive corpora, which can benefit downstream tasks but also perpetuate the formal-domain bias and embed cultural or social prejudices (Bordia & Bowman, 2019). Supervised fine-tuning on annotated narrative datasets has been shown to substantially improve valence accuracy by aligning model parameters with the target register's features (Demszky, 2023). However, this process demands sizeable, high-quality annotations of Flemish narratives, data that are currently scarce. Furthermore, LLMs risk amplifying biases in sentiment judgments, differentiating affective language based on gendered or demographic stereotypes unless rigorous bias auditing and inclusive data curation practices are enforced (Bordia & Bowman, 2019). Future research should prioritize the creation of standardized

Flemish narrative valence benchmarks based on manually annotated corpora of everyday Flemish texts that enable systematic evaluation of LLM-based approaches (Augustyniak et al., 2023) which have the potential to bridge computational and psychological research. This urgency is amplified in mental health research, where daily narratives, critical for capturing nuanced emotional states, remain underrepresented in NLP studies. Until LLMs overcome domain mismatches and achieve comparable robustness, lexicon tools remain indispensable for scalable, ecologically valid valence analysis in Flemish.

## 8. Conclusion

This study demonstrates that lexicon-based tools like Pattern.nl remain the gold standard for valence analysis in low-resource Flemish, outperforming Dutch-tuned LLMs in coverage and fairness despite their contextual limitations. Moving forward, establishing standardized benchmarks for Flemish valence analysis, grounded in manually annotated, ecologically valid narratives will enable targeted improvements in data augmentation, bias mitigation, and hybrid modeling. Simultaneously, the call for personalized LLMs (Nature Machine Intelligence, 2024) highlights a critical tension: while LLMs cannot replicate human empathy, their integration with lexicon-driven frameworks may bridge the gap between scalable automation and culturally grounded emotion research. Until such synergies are realized, lexicon tools will continue to anchor valence analysis in low-resource contexts, ensuring both fairness and methodological rigor.

## 9. Limitations

While this study advances understanding of valence analysis in Flemish, its scope is constrained by several factors. First, Dutch and Flemish remain critically under-represented in both training data and available LLMs, reflecting broader systemic gaps in low-resource language NLP. Though recent efforts (Mattheus, 2024) aim to bridge this gap, developing state-of-the-art systems for Flemish is hampered by sparse language resources, limited computational infrastructure, and a shortage of technically proficient users, challenges endemic to under-resourced languages. Our evaluation of three Dutch-tuned LLMs, while informative does not account for larger models (e.g., LLaMA-2 70B), newer architectures or hybrid approaches that may outperform current benchmarks. Second, while our corpus of ~25,000 narratives provides rich ecological validity, its reliance on a single prompt structure risks conflating valence expression with elicitation method, limiting generalizability to other narrative contexts. Finally, while lexicon tools circumvent LLMs' technical limitations, they lack contextual nuance (e.g., resolving sarcasm or cultural idioms) and cannot replicate human empathy, a gap underscored by critiques of LLMs' inability to interpret emotional subtext (Nature Machine Intelligence, 2024).

**Supplementary Material**

## A. Data Preparation

Participants were recruited from the community via flyers, online advertisements, and word-of-mouth referrals. Eligibility criteria included being native Dutch speakers, residing in Belgium, being at least 18 years of age, and owning a fully functional smartphone. Interested individuals completed an initial eligibility survey, with eligibility criteria verified during a subsequent online introduction session.

Participants received compensation of up to €250 for participating in a 70-day (10-week) experience sampling protocol, accompanied by biweekly online surveys. The breakdown included €0.50 per completed experience sampling prompt (maximum of 280 prompts or €140 total), €10 for each short survey at 2, 4, 6, and 8 weeks (maximum €40), and €15 for each long survey at baseline and 10 weeks (maximum €30). An additional bonus of €40 was given to participants completing at least 60 days. Payment was issued in full upon study completion. Continued participation required completion of at least 75\% of prompts, with verbal responses having a minimum length of 25 words. Regular compliance checks and biweekly summary reports were provided to participants, with reminders issued as necessary.

Initially, 115 participants enrolled (age range: 18–65, M = 27.26, SD = 9.86; gender: 58 female, 56 male, 1 other). Of these, 10 voluntarily withdrew, and 3 were dismissed due to poor compliance (response rates below 50\%). Thus, 102 participants completed the study (age range: 18–65, M = 26.47, SD = 8.87; 52 women, 49 men, 1 other).

Ethical approval was obtained from the KU Leuven Social and Societal Ethics Committee (SMEC), protocol G-2023-6379-R3(AMD). Data collection spanned from August 2023 to July 2024. All study materials and instructions were administered in Dutch.

### A.1 Procedure and Materials

Participants engaged in a 70-day experience sampling protocol via a dedicated mobile application (m-Path; Mestdagh et al., 2023), receiving four prompts daily at pseudorandom intervals between 9 AM and 9 PM, spaced at least one hour apart. At each prompt, participants responded to "What is going on now or since the last prompt, and how do you feel about it?" Responses were typically recorded as 1-minute voice messages but could optionally be typed (3-4 sentences). Participants then rated their current emotional valence on a slider scale ranging from -50 (very unpleasant) to +50 (very pleasant). Simultaneously, m-Path recorded sensor data (GPS coordinates, ambient noise, step counts, and app usage for Android devices).

All assessment tools were validated previously in Dutch-speaking samples. Participants completed additional online surveys after 2, 4, 6, and 8 weeks, and a comprehensive survey at 10 weeks, which included validation questions about recent visited locations.

Participants received biweekly compliance summary reports via email, detailing prompt completion rates, description length adequacy, modality (voice vs. text), valence ratings, linguistic content trends (using LIWC 2015 Dutch translation), and accumulated compensation. These reports also indicated whether participants earned bonuses and their overall compensation.

Voice-recorded responses were automatically transcribed using a proprietary algorithm developed by the Department of Electrical Engineering (ESAT) at KU Leuven (Tamm et al., 2024). Transcripts and typed responses were integrated into a unified data file, aligned manually with online survey data into five study intervals: days 1–14, 15–28, 29–42, 43–56, and 57–70. Missing survey data were noted where applicable.

### A.1.2 Analytic Tools and Experimental Setup

Textual responses were analyzed using Linguistic Inquiry and Word Count (LIWC) with its Dutch lexicon, generating category-specific linguistic scores for each response. Additionally, the Pattern.nl sentiment analysis algorithm provided continuous valence estimates from -1 (very negative) to +1 (very positive), accounting for word context and grammatical function. Model evaluations were performed using a NVIDIA RTX-5000 GPU and implemented in PyTorch, with models obtained via Hugging Face APIs.

### A.2 Single Participant (Pilot Study)

We conducted a preliminary pilot study involving one participant's data from our dataset to identify the best performing models. We also compared the performances of these models with English prompt and Dutch prompt.

**English Prompt:** "You are a Dutch language expert analyzing the valence of Belgian Dutch texts. Participants responded to: "What is going on now or since the last prompt, and how do you feel about it?"

Carefully read the response of the participant: "text".

Your task is to rate its sentiment from 1 (very negative) to 7 (very positive).

Return ONLY a single numerical rating enclosed in brackets, (e.g. [X]), with no additional text, explanations, or formatting.

Output Format: [number] .

Replace "number" with the integer score (1-7)."

**Dutch Prompt:** "Je bent een Nederlandse taalexpert die de valentie van Belgisch Nederlandse teksten analyseert. Deelnemers reageerden op:

"Wat speelt er nu of sinds de vorige beep en hoe voel je je daarover?". Lees zorgvuldig het antwoord van de deelnemer: "text".

Het is jouw taak om het sentiment te beoordelen van 1 (zeer slecht) tot 7 (zeer goed). Geef ALLEEN een cijfer tussen haakjes (bijv. [X]), zonder extra tekst, uitleg of opmaak. Niet uitleggen. Uitvoerformaat: [getal]. Vervang "getal" door de gehele score (1-7)."

As illustrated in Table 1, we selected ChocoLlama-8B-Instruct, GEITje-7B-ultra, Reynaerde-7B-Chat, and Llama 2-7B. We also ran the dataset with LIWC15 and Pattern. The models returned values for all the texts. Among these LLMs, when prompted in English, ChocoLlama-8B-Instruct achieved the highest polyserial correlation (r = 0.55) with the user's ratings followed by GEITje-7B-ultra (r = 0.42) and Reynaerde-7B-Chat (r = 0.33). Surprisingly, Llama-2-7B returned no values for all the texts when prompted in English but performed better than GEITje-7B-ultra (r = 0.0002) and Reynaerde-7B-Chat (r = 0.42). To maintain uniformity, we decided to proceed with only English prompt for further experiments. Between LIWC15 and Pattern.nl, LIWC15 showed strong negative (negemo: r = -0.54) and positive (posemo: r = 0.41) correlations, and Pattern.nl showed positive correlation with the user's ratings (r = 0.44).

Table 1. Correlation coefficients across models and prompts

| Model (variable) | Prompt | Pearson *r* | Polyserial *r* |
|---|---|---|---|
| LIWC15 (posemo) | - | 0.41 | - |
| LIWC15 (negemo) | - | -0.54 | - |
| Pattern.nl | - | 0.44 | - |
| ChocoLlama-8B-Instruct | English | 0.47 | 0.55 |
| ChocoLlama-8B-Instruct | Dutch | 0.18 | 0.27 |
| Llama-2-7B | English | N/A | N/A |
| Llama-2-7B | Dutch | 0.37 | 0.46 |
| GEITje-7B-ultra | English | 0.33 | 0.42 |
| GEITje-7B-ultra | Dutch | 0.0001 | 0.0002 |
| Reynaerde-7B-Chat | English | -0.1 | 0.33 |
| Reynaerde-7B-Chat | Dutch | 0.2 | 0.42 |

In the following sections, we present the additional experiments conducted and the results observed for the entire dataset using the English prompt:

## A.3 Distribution of Ratings (All Participants/Users, LIWC, Pattern.nl)

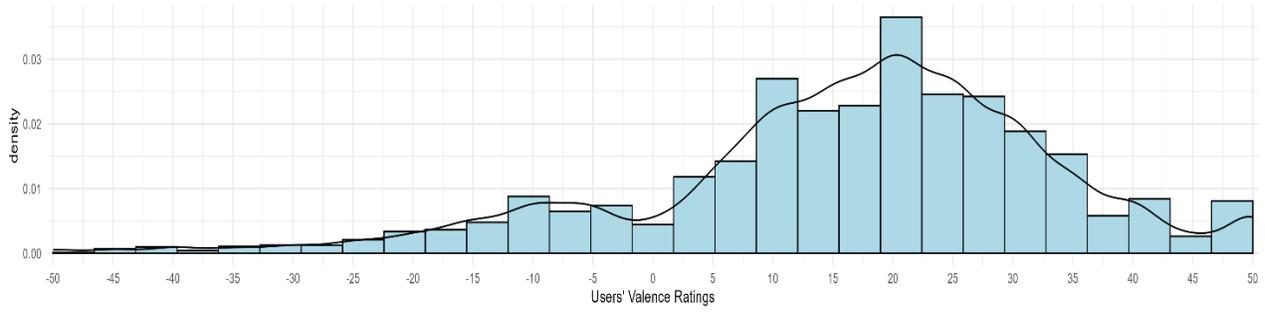

Fig 1. Distribution of users' self-reported valence ratings

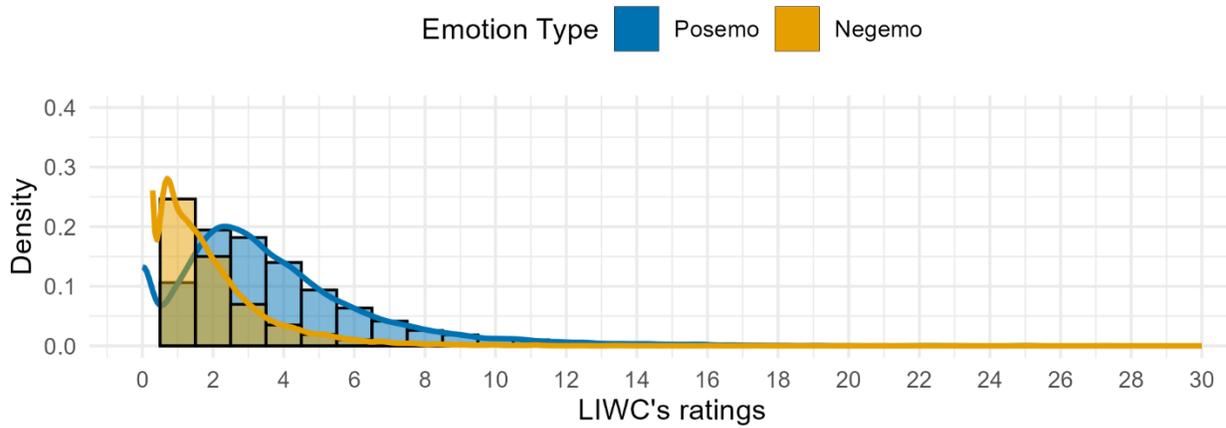

Fig 2. Distribution of LIWC's scores

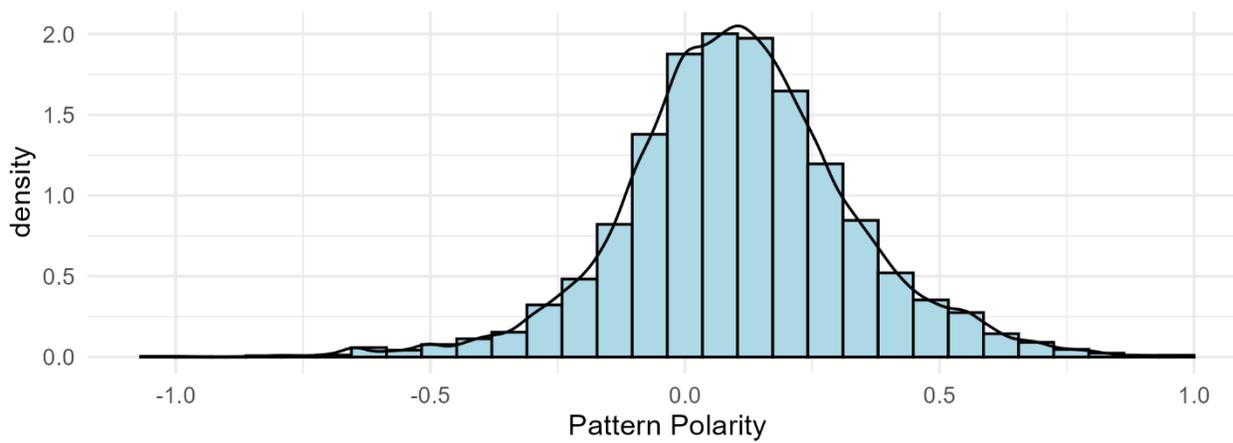

Fig 3. Distribution of Pattern.nl's polarity

## A.4 Zero-shot setting results (All participants)

In the zero-shot condition, as discussed previously in the Results section, the models did not return values for all the texts, unlike LIWC and Pattern. ChocoLlama-8B-Instruct returned values for only 17378 texts, GEITje-7B-ultra for 9445 texts, and Reynaerde-7B-Chat for 446 estimates only. Pattern.nl turned out to be performing better on most of the texts than these models.

Polyserial correlation analyses revealed significant relationships between user ratings and the outputs of all three language models tested, with all user-model comparisons yielding p-values < 0.001. However, correlations between Reynaerde and traditional sentiment tools varied: LIWC15 (posemo) showed no significant association (p = 0.12), while LIWC15 (negemo) approached significance (p < 0.05) but did not meet the threshold for statistical reliability. In contrast, Pattern.nl demonstrated a modest yet statistically significant correlation with Reynaerde (p = 0.022). GEITje showed strong, statistically significant correlations with all lexicon-based models—LIWC15 (posemo and negemo) and Pattern.nl (all p < 0.001), indicating robust agreement. Similarly, ChocoLlama's outputs correlated significantly with LIWC15 and Pattern.nl (all p < 0.001).

### A.4.1 Distributional Analysis of the Models' Ratings

ChocoLlama-8B-Instruct showed a relatively balanced distribution compared to the other two LLMs (peaking around 3 and 5). GEITje-7B-ultra's ratings were skewed towards scores around 5. Reynaerde-7B-Chat's predictions had the narrowest distribution amongst all these three models.

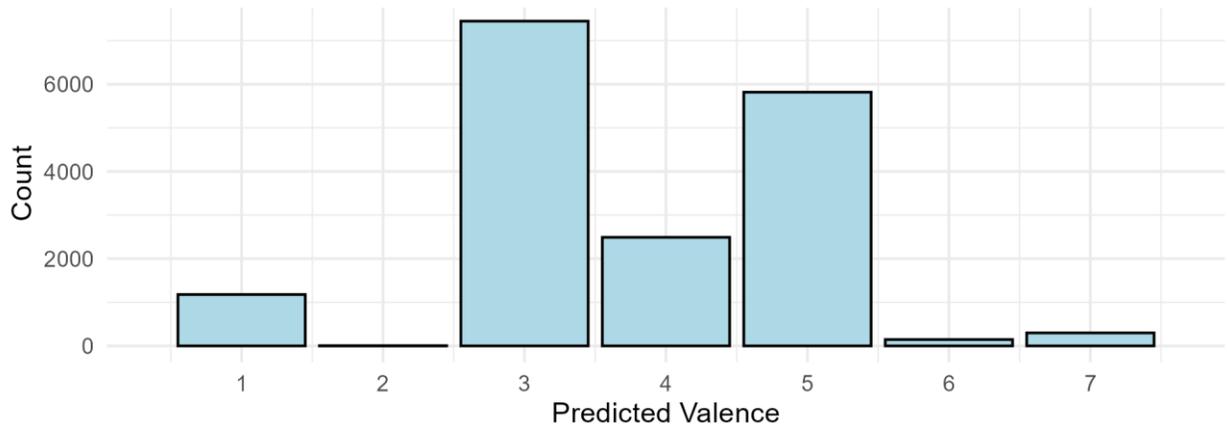

Fig 4. Distribution of Chocollama-8B-instruct's predicted valence scores

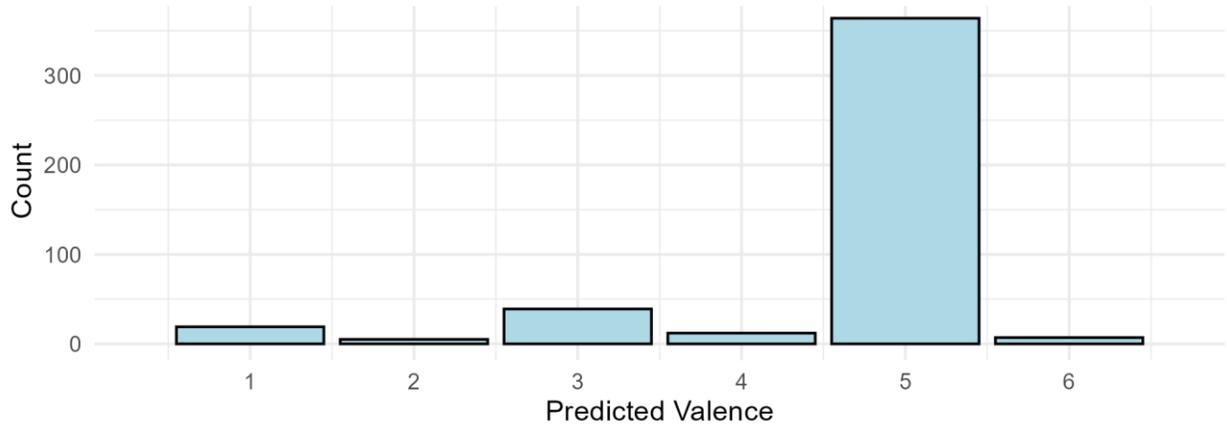

Fig 5. Distribution of Reynaerde-7B's valence scores

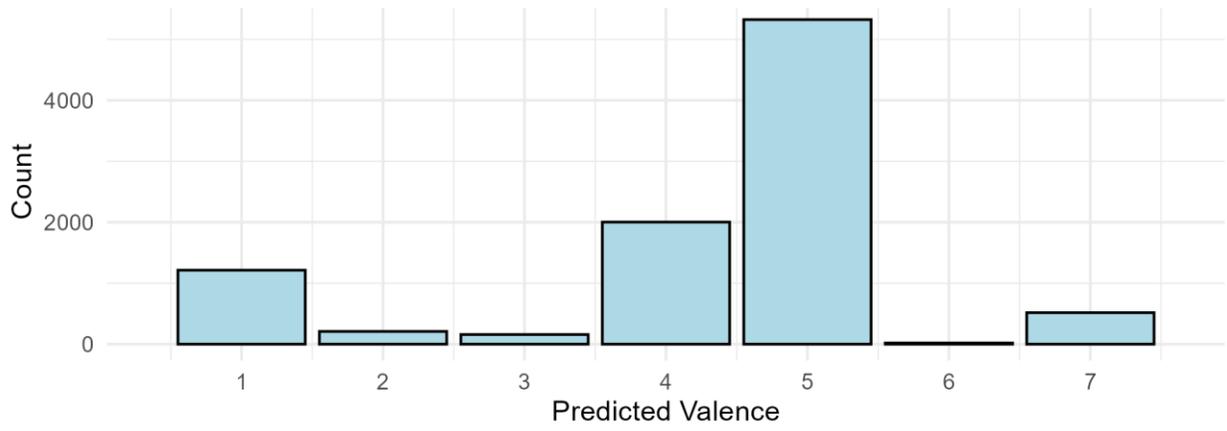

Fig 6. Distribution of GEITje-7B's valence scores

## A.4.2 Box Plot Analysis of Users' Ratings vs. Model's Predictions

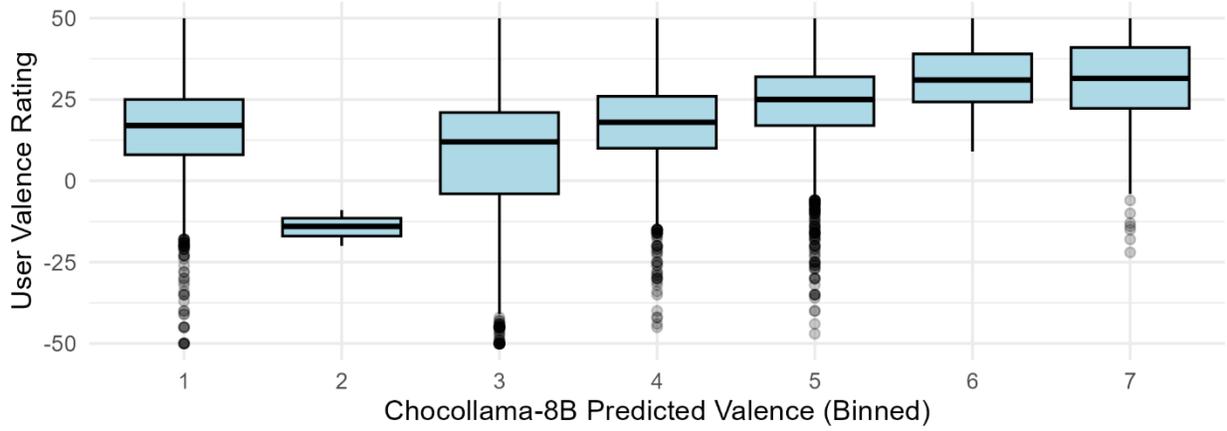

Fig 7. Chocollama-8B-instruct

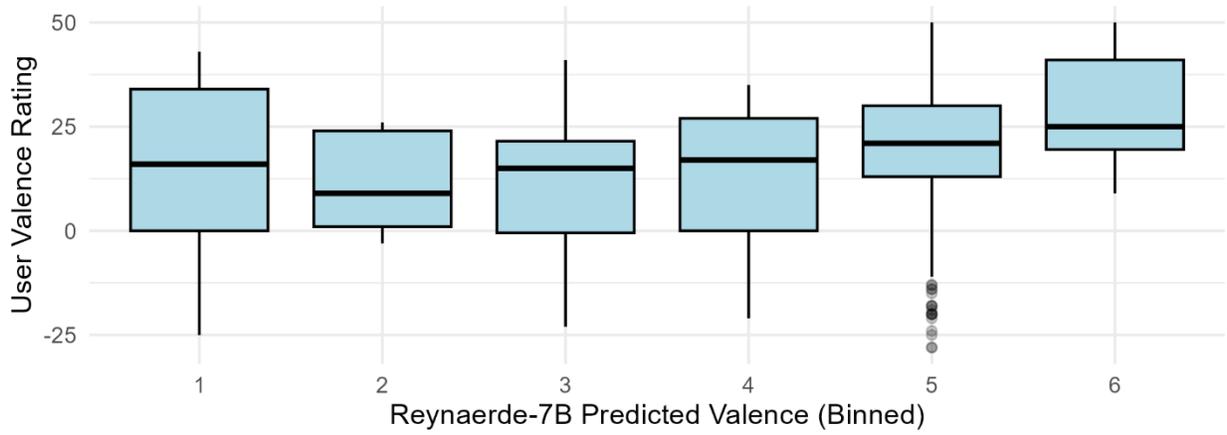

Fig 8. Reynaerde-7B

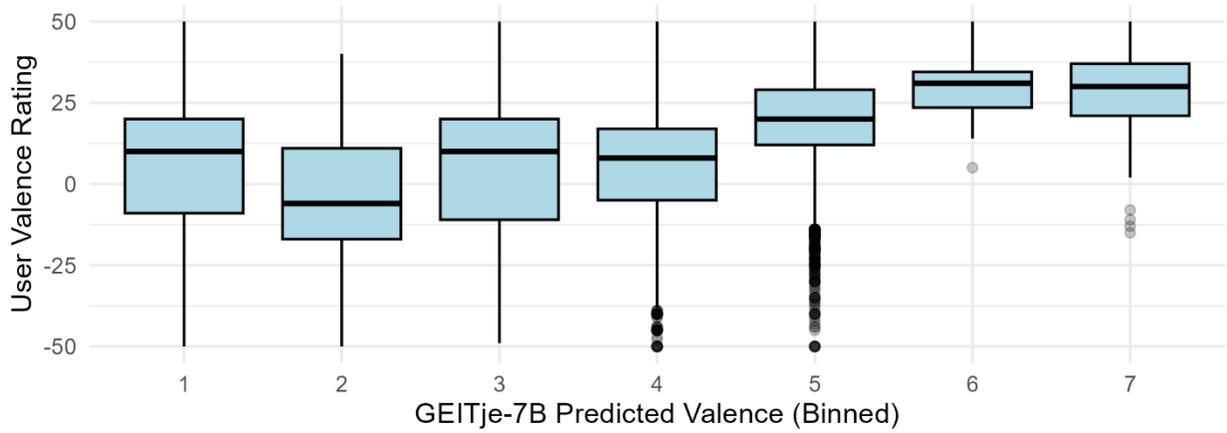

Fig 9. GEITje-7B

## A.4.3 Scatter Plots

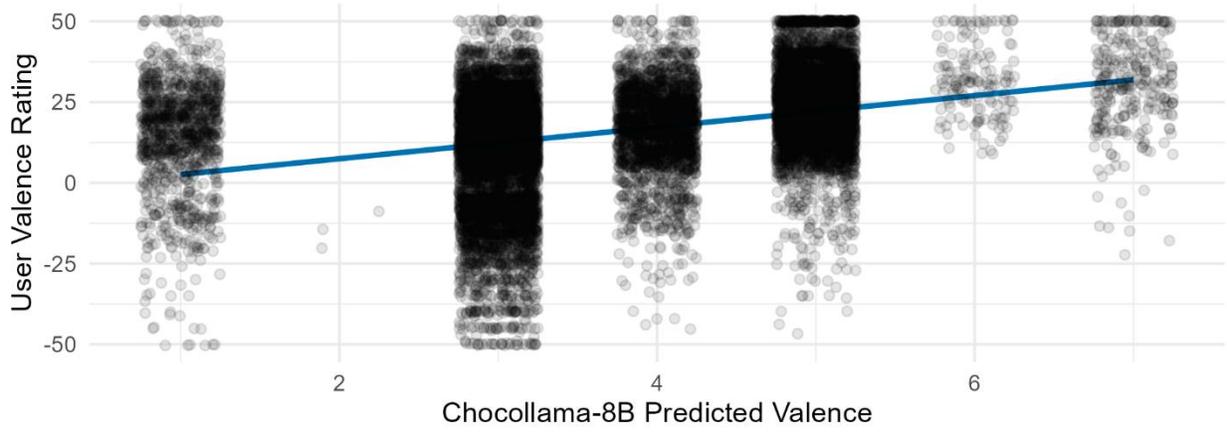

Fig 10. Chocollama-8B-instruct

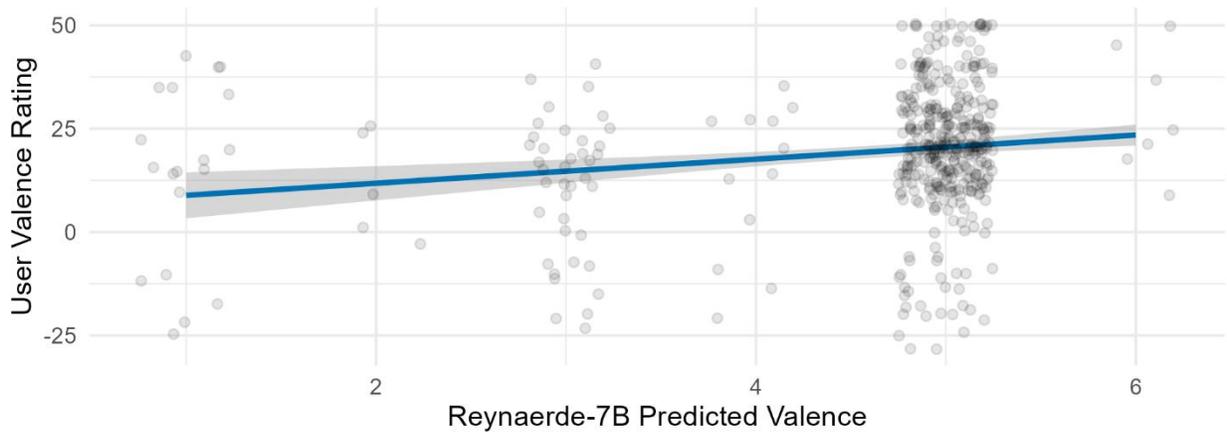

Fig 11. Reynaerde-7B

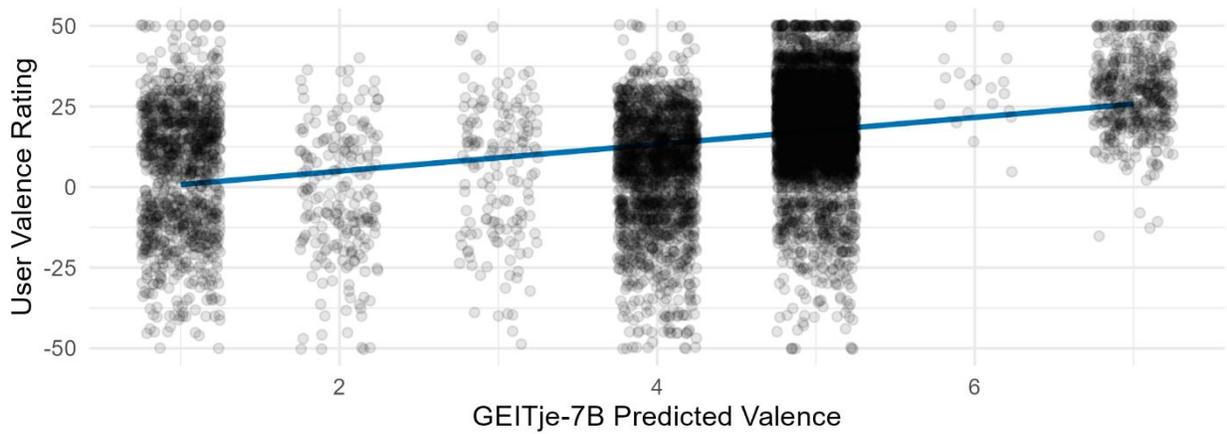

Fig 12. GEITje-7B

### A.5 Few-shot setting results (All participants)

We also attempted a few-shot setting by providing five examples from the text along with the user ratings.

### A.5.1 Prompt used

In this prompt, "(Text)" has been replaced with the actual responses of the participants.

> "You are a Dutch language expert analyzing the valence of Belgian Dutch texts. Participants responded to the question:
> "What is going on now or since the last prompt, and how do you feel about it?"
> They also rated their own emotional valence on a continuous scale from -50 (very negative) to +50 (very positive). Your task is to read each response: {text} and rate its sentiment from 1 (very negative) to 7 (very positive). Return ONLY a single numerical rating enclosed in square brackets, e.g. [X]. Provide no explanation or additional text.
> Output Format: [number].
>
> Below are a few examples of an input and the participant's valence rating to guide your rating:
>
> Input: (Text1)
> Output: [10]
>
> Input: (Text2)
> Output: [-10]
>
> Input: (Text3)
> Output: [30]
>
> Input: (Text4)
> Output: [45]
>
> Input: (Text5)
> Output: [40]"

We observed a similar trend in the LLMs chosen returned values for only a portion of the texts. Pattern.nl proved to be more effective than LIWC.

Table 2. Few-shot coverage and correlation coefficients across models and benchmarks wrt users

| Model | Coverage (N/%) | Pearson *r* | Polyserial *r* |
|---|---|---|---|
| LIWC15 (posemo) | 24,848 / 99.9% | 0.21 | 0.23 |
| LIWC 15 (negemo) | 24, 848 / 99.9% | -0.23 | -0.26 |
| Pattern.nl | 24,848 / 99.9% | 0.31 | 0.33 |
| Reynaerde-7B-Chat (English prompt) | 7,219 / 29.0% | 0.03 | 0.036 |
| GEITje-7B-ultra (English Prompt) | 11,323 / 45.6% | -0.08 | -0.09 |
| Chocollama-8B-instruct (English prompt) | 5,266 / 21.2 % | -0.03 | -0.029 |

In the few-shot setting, we provided five example texts with user ratings. Significance values (p-values) from polyserial correlation analyses were as follows:

### A.5.2 Polyserial Correlation Between the models

Table 3. Polyserial correlation and significance for model-model comparison

| Comparison | r | p | Significance |
|---|---|---|---|
| LIWC15 (posemo) vs Reynaerde-7B-Chat | 0.11 | < 0.001 | significant |
| LIWC 15 (negemo) vs Reynaerde-7B-Chat | -0.004 | 0.916 | non-significant |
| Pattern vs Reynaerde-7B-Chat | 0.044 | 0.001 | significant |
| | | | |
| LIWC15 (posemo) vs GEITje-7B-ultra | 0.068 | 0.003 | significant |
| LIWC 15 (negemo) vs GEITje-7B-ultra | 0.131 | <0.001 | significant |
| Pattern vs GEITje-7B-ultra | -0.05 | 0.7 | non-significant |
| | | | |
| LIWC15 (posemo) vs Chocollama-8B-instruct | -0.052 | 0.479 | non-significant |
| LIWC 15 (negemo) vs Chocollama-8B-instruct | -0.004 | 0.001 | significant |
| Pattern vs Chocollama-8B-instruct | -0.046 | 0.940 | non-significant |

### A.5.3 Distributional Analysis of the Models' Ratings

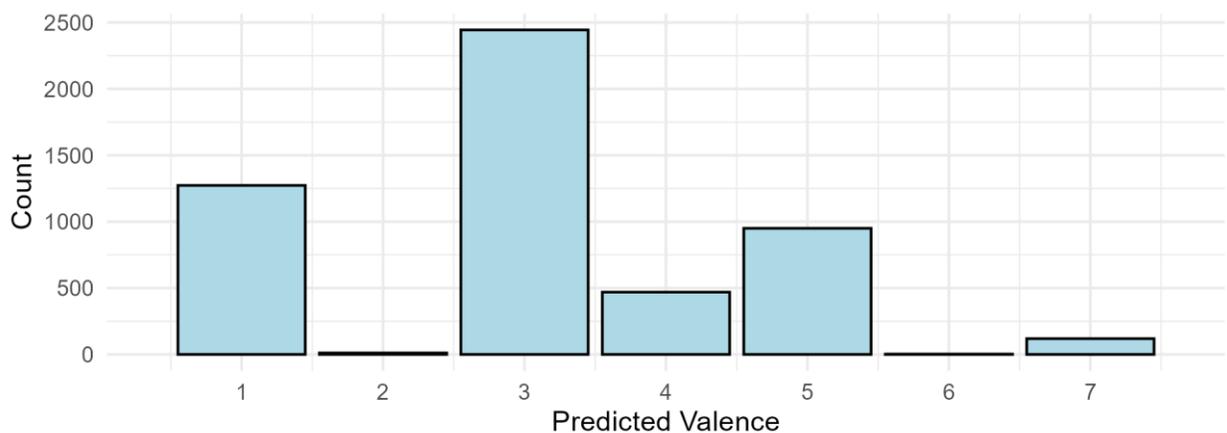

Fig 13. Distribution of Chocollama-8B-instruct's predicted valence scores

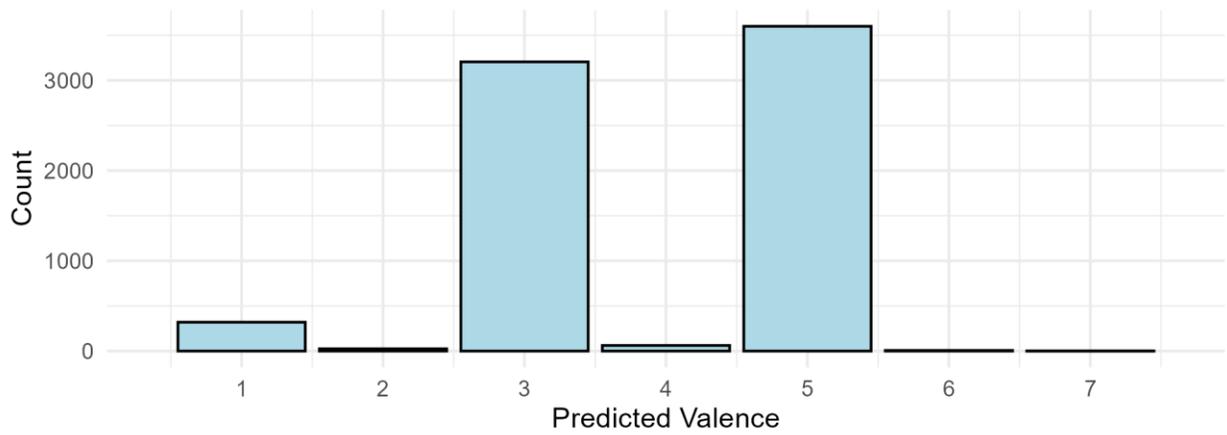

Fig 14. Distribution of Reynaerde-7B's valence scores

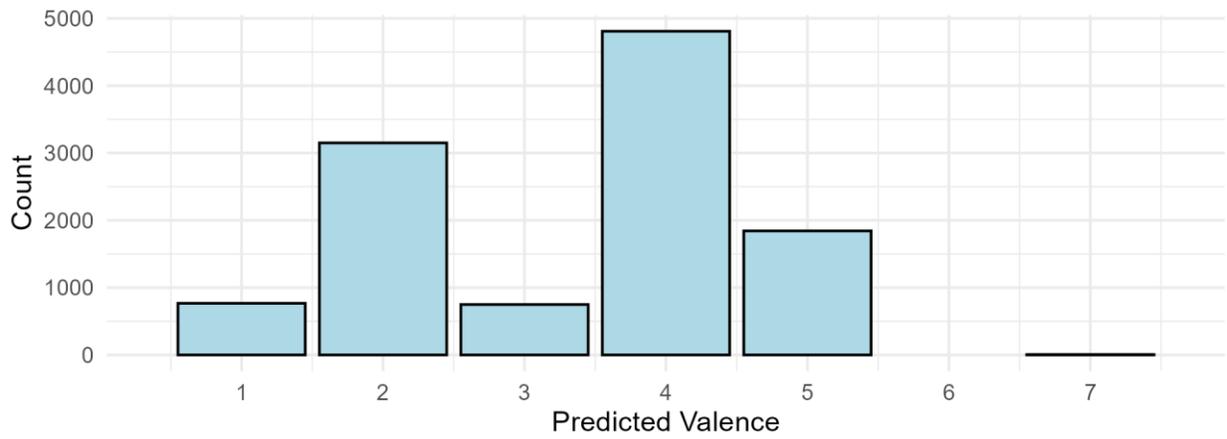

Fig 15. Distribution of GEITje-7B's valence scores

**A.5.4 Box Plots of Users' Ratings vs. Model's Predictions**

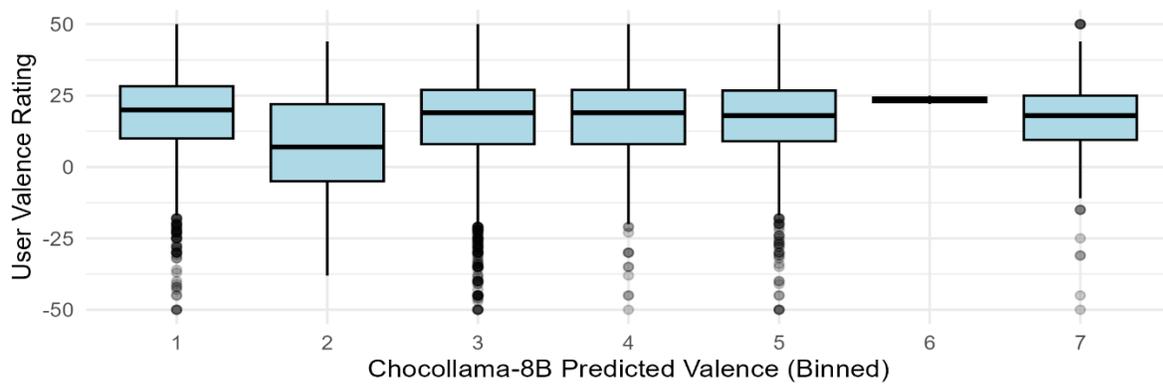

Fig 16. Chocollama-8B-instruct

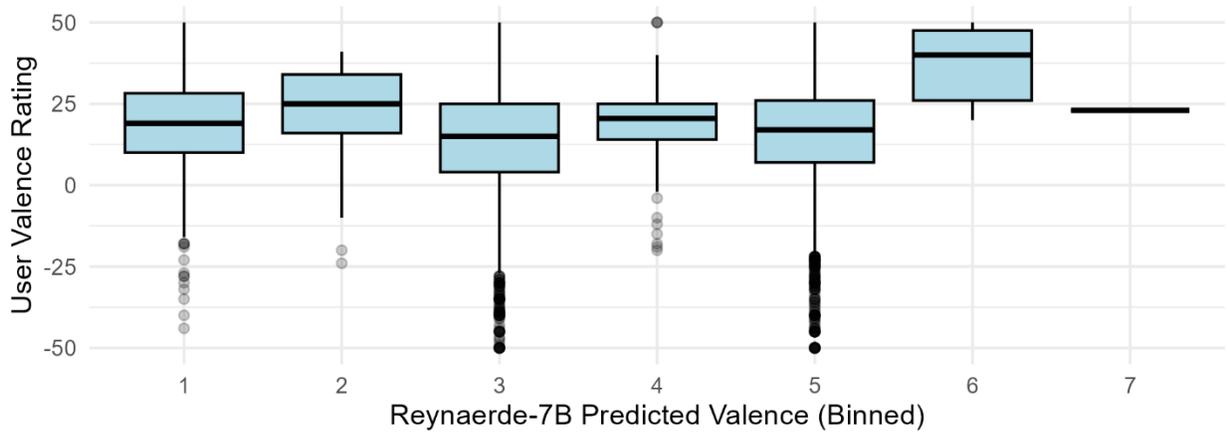

Fig 17. Reynaerde-7B

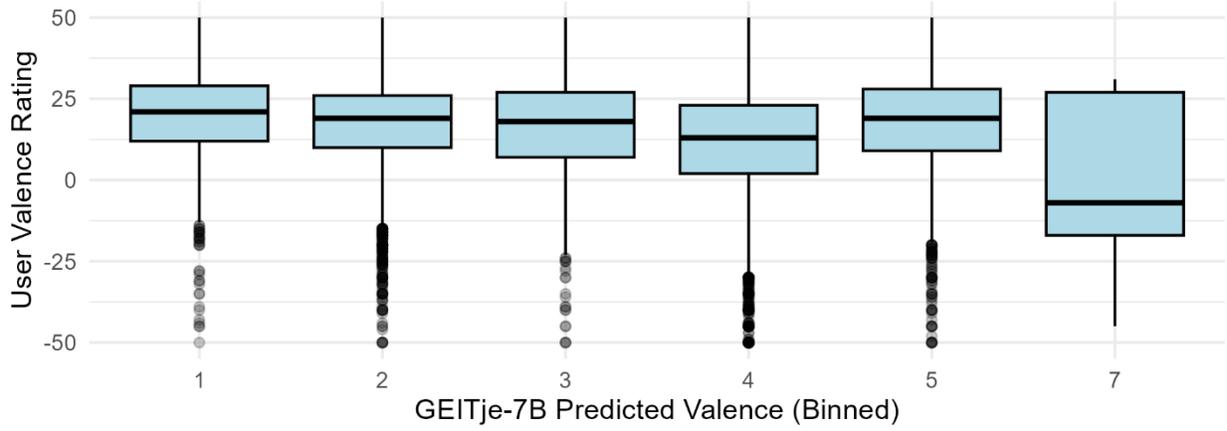

Fig 18. GEITje-7B

### A.5.5 Scatter Plots

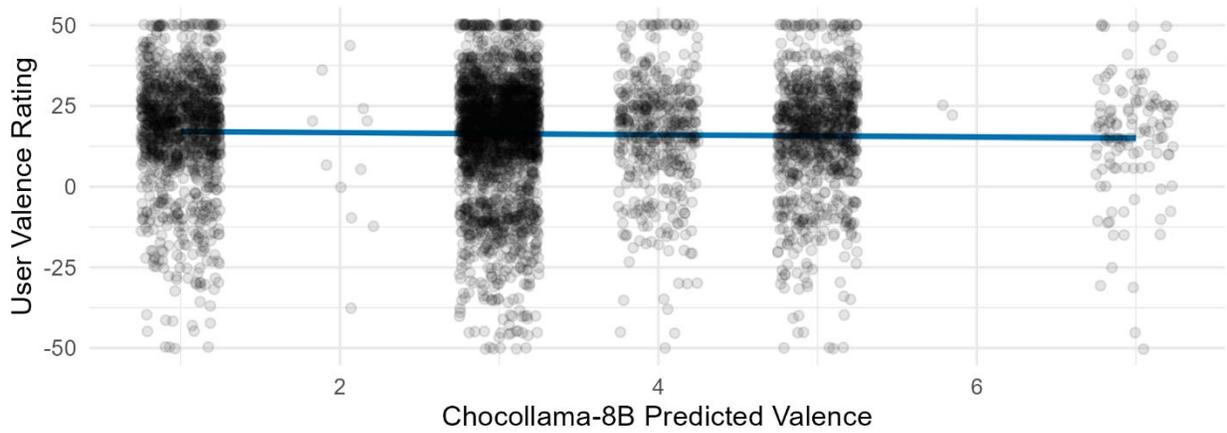

Fig 19. Chocollama-8B-instruct

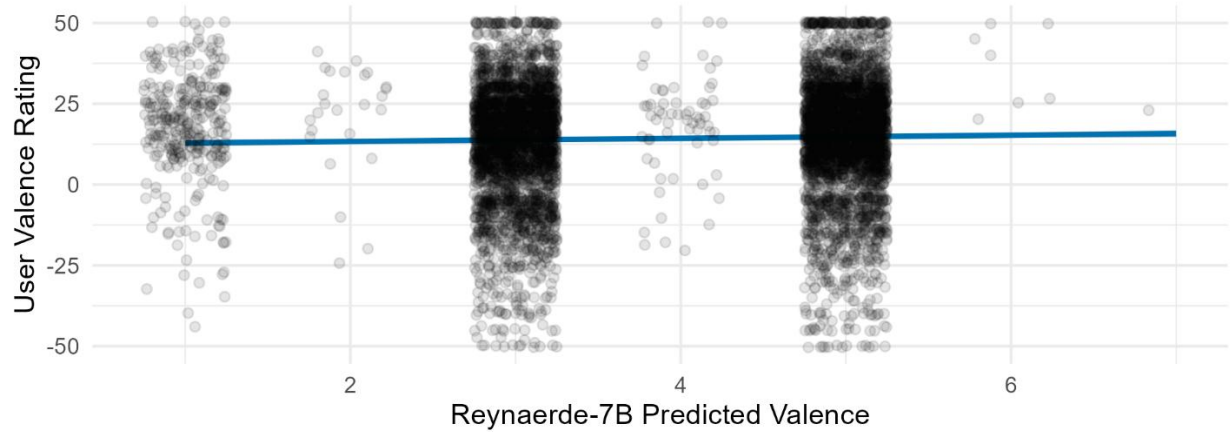

Fig 20. Reynaerde-7B

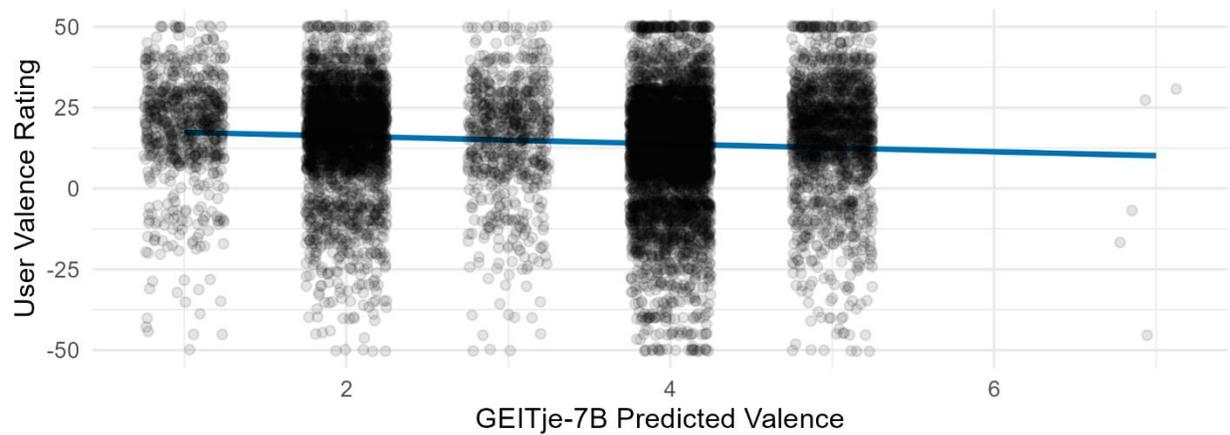

Fig 21. GEITje-7B

## A.6 References